\documentclass[conference]{IEEEtran}
\usepackage{times}

\usepackage[numbers]{natbib}
\usepackage{multicol}
\usepackage[bookmarks=true]{hyperref}

\usepackage{mathtools}
\usepackage{amsfonts}  
\usepackage{graphics}
\usepackage[pdftex]{graphicx}
\usepackage{wrapfig}
\usepackage{caption}
\usepackage{color}
\usepackage{dsfont}
\usepackage[]{mdframed}
\usepackage{algorithm}
\usepackage{algorithmic}
\usepackage{xparse}
\usepackage{amsmath}
\usepackage{mathtools}
\usepackage{amssymb}
\usepackage{amsthm}
\usepackage{booktabs}
\usepackage{xspace}
\usepackage{cleveref}
\usepackage{xcolor}
\usepackage{fontawesome}
\usepackage{romannum}
\usepackage{bm}

\definecolor{MyDarkGreen}{rgb}{0.02,0.6,0.02}

\DeclarePairedDelimiterX{\infdivx}[2]{(}{)}{%
  #1\;\delimsize\|\;#2%
}

\usepackage{expl3}
\ExplSyntaxOn
\newcommand\latinabbrev[1]{
  \peek_meaning:NTF . {
    #1\@}%
  { \peek_catcode:NTF a {
      #1.\@ }%
    {#1.\@}}}
\ExplSyntaxOff

\NewDocumentCommand \proposition {g g g g} {\texttt{#1}(#2
  \IfValueTF{#3}{,\,#3}{}
  \IfValueTF{#4}{,\,#4}{}
  )
}

\NewDocumentCommand \actioncall {g g g g} {\text{#1}(#2
  \IfValueTF{#3}{,#3}{}
  \IfValueTF{#4}{,#4}{}
  \texttt{)}
}

\def \MethodName {Prompting with the Future\xspace}
\def \MethodAcronym {PWTF\xspace}

\usepackage{animate}
\newcommand{\IfDefinedSwitch}[3]{%
  \ifdefined#1
    #2 
  \else
    #3 
  \fi
}
\newcommand{\embedVideo}{Embed in the wild generation video (any placeholder text)}

\begin{document}

\title{Prompting with the Future: Open-World Model Predictive Control with Interactive Digital Twins}

\author{
    Chuanruo Ning \quad 
    Kuan Fang$^{\dagger}$ \quad 
    Wei-Chiu Ma$^{\dagger}$ \quad 
    \vspace{0.5mm} \\
    Cornell University \vspace{0.5mm} \\
    \url{https://prompting-with-the-future.github.io/}  \vspace{-3mm}
    }


\twocolumn[{%
    \renewcommand\twocolumn[1][]{#1}%
    \maketitle
    
 \begin{center}
     \begin{minipage}[t]{0.769\textwidth}
        \includegraphics[width=\textwidth]{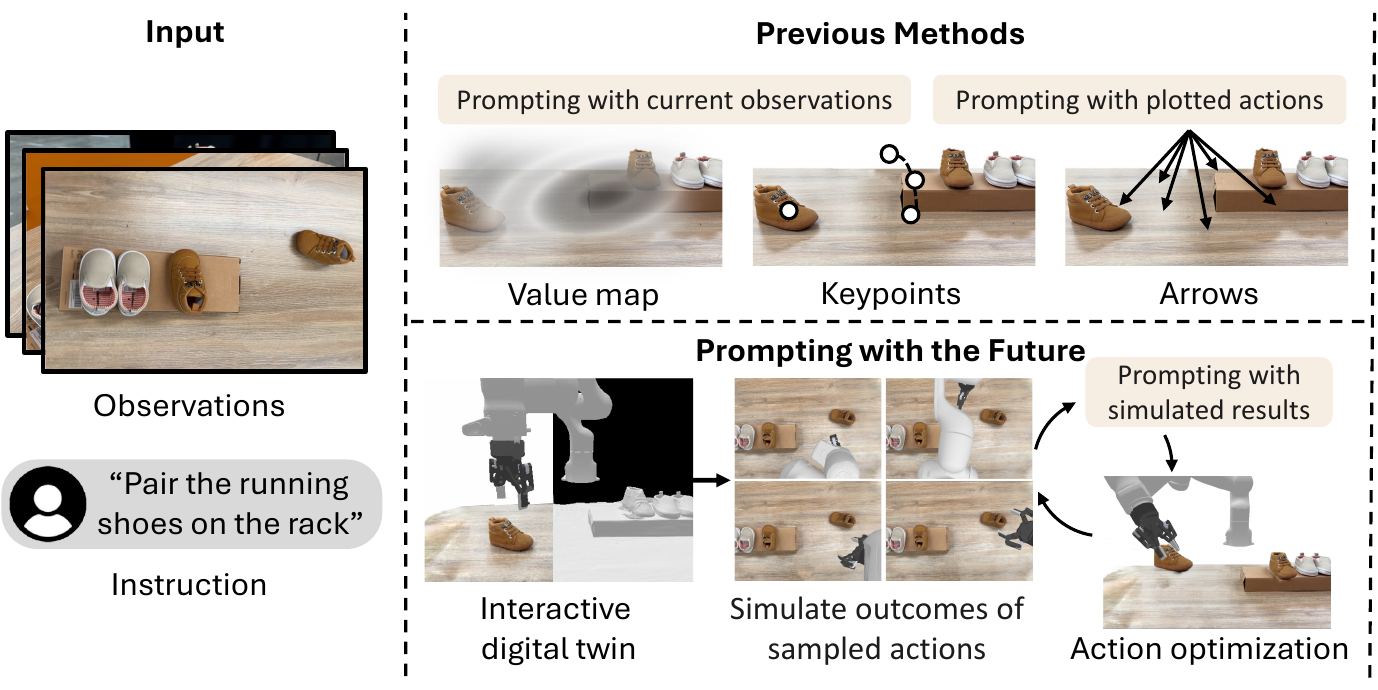}
     \end{minipage}
     \hfill
     \begin{minipage}[t]{0.212\textwidth}
        \centering
        \IfDefinedSwitch{\embedVideo}{
        \animategraphics[autoplay,loop,width=\linewidth, trim={0 0 0 0}, clip]{5}{figures/teaser_frames/}{000}{89}
        }
        {\includegraphics[width=\linewidth]{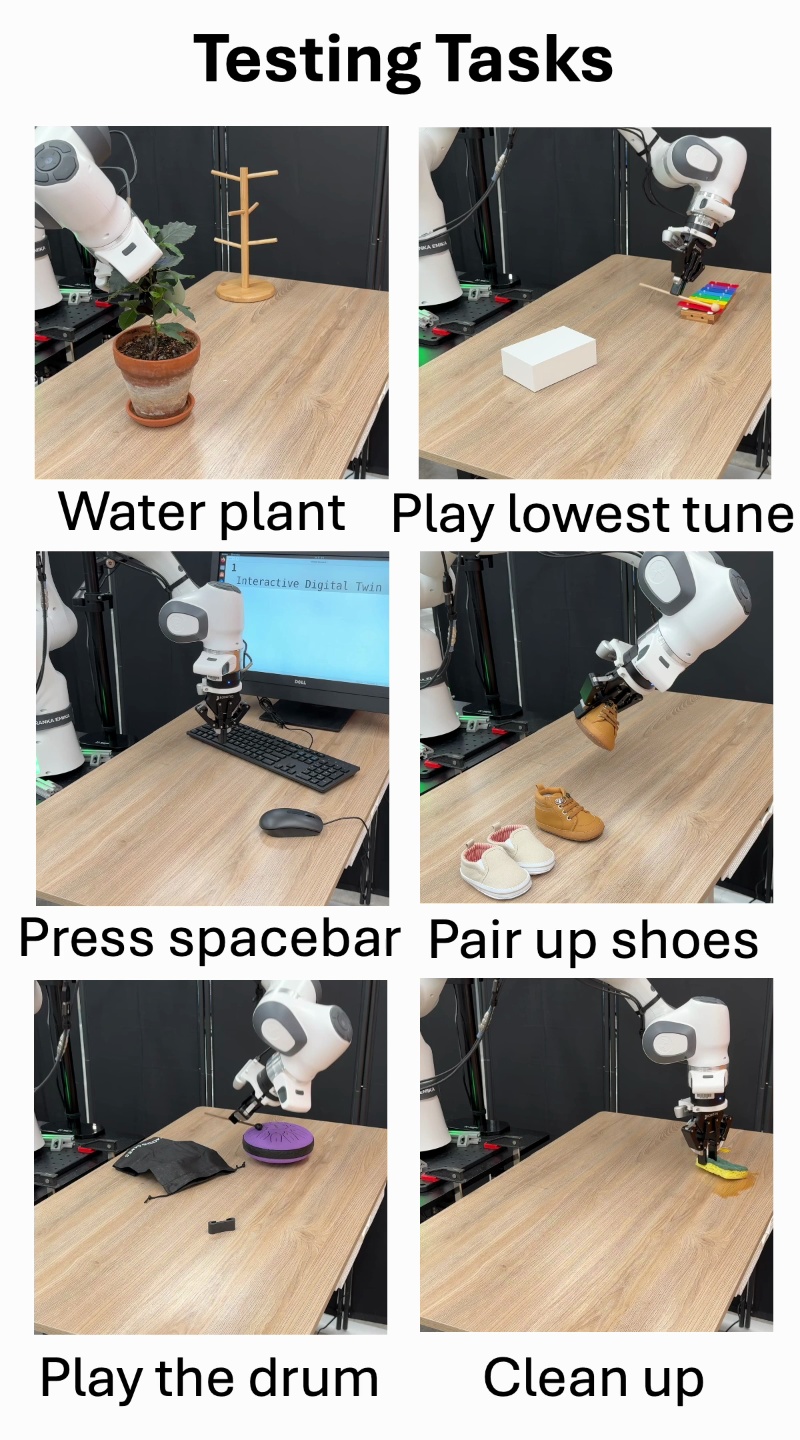}
        }
     \end{minipage}
     \vspace{-2.5mm}
    \captionof{figure}{
    \textbf{Prompting with the Future.} 
    We enable VLM-driven motion planning conditioned on free-form instructions by prompting with simulated future outcomes generated from an interactive digital twin.
In contrast, previous methods~\cite{huang2023voxposer, fang2024moka, nasiriany2024pivot} prompt VLMs with current observations and hand-defined action primitives, demanding implicit physical reasoning that VLMs often struggle to perform reliably.
    Please use \emph{Adobe Acrobat Reader} to view embedded video demonstrations.
    }
    \label{fig:teaser}
    \end{center}
    \vspace{3.5mm}
}]

\begin{abstract}
Open-world robotic manipulation requires robots to perform novel tasks described by free-form language in unstructured settings. While vision–language models (VLMs) offer strong high-level semantic reasoning, they lack the fine-grained physical insight needed for precise low-level control. To address this gap, we introduce \MethodName (\MethodAcronym), a model-predictive control framework that augments VLM-based policies with explicit physics modeling. \MethodAcronym builds an interactive digital twin of the workspace from a quick handheld video scan, enabling prediction of future states under candidate action sequences. Instead of asking the VLM to predict actions or results by reasoning dynamics, the framework simulates diverse possible outcomes, renders them as visual prompts with adaptively selected camera viewpoints that expose the most informative physical context. A sampling-based planner then selects the action sequence that the VLM rates as best aligned with the task objective. We validate \MethodAcronym on eight real-world manipulation tasks involving contact-rich interaction, object reorientation, and clutter removal, demonstrating significantly higher success rates than state-of-the-art VLM-based control methods. Through ablation studies, we further analyze the performance and demonstrate that explicitly modeling physics, while still leveraging VLM semantic strengths, is essential for robust manipulation.
\end{abstract}

\footnotetext{$^{\dagger}$ Equal advising.}

\section{Introduction}
\label{sec:introduction}
General-purpose robots must perform novel, human-specified tasks in unstructured, highly varied settings. Recent vision–language models (VLMs) have made impressive strides in high-level semantic reasoning and zero-shot visual question answering \cite{ahn2022can, duan2024manipulate, huang2023voxposer}. Yet their understanding of fine-grained physical interactions typically cannot support the precise, low-level control required in real-world manipulation. Take the seemingly simple job of tidying a cluttered shoe rack (\Cref{fig:teaser}). The robot needs to, detect and identify each shoe, infer the desired final arrangement, and plan and execute gripper motions that achieve accurate translations and rotations while avoiding collisions. Achieving open-world manipulation thus hinges on seamlessly fusing semantic reasoning with the fine-grained, physics-aware control that real-world tasks demand.

Recent works have investigated various ways to employ VLMs for robotic control.
Fine-tuned on massive demonstration trajectories through imitation learning, VLMs can serve as the backbone of policies to ground language instructions in visual observations~\cite{brohan2023rt, kim2024openvla, o2023open}. 
However, robotics datasets remain orders of magnitude smaller than the question-answering datasets used to train VLMs, limiting the generalization of learned policies to novel environments and tasks. Alternatively, an increasing number of works propose to employ VLMs in zero-shot settings either with the observed images~\cite{huang2023voxposer} or sampled actions overlaid on the images as visual prompts ~\cite{fang2024moka, nasiriany2024pivot}. 
While these approaches show promise in simple tasks, they require VLM to infer future interactions, thus struggling with more complex scenarios involving intricate contact, dynamics, and motion due to insufficient physical understanding.

In this work, we propose a fundamentally different approach to address this challenge by decoupling semantic understanding from physical reasoning.
In contrast to prior works, which require VLM to implicitly reason dynamics, we employ interactive digital twins of the real-world environment to complement the VLM's capabilities. 
Through physical simulation, the digital twin explicitly models the dynamics of the manipulation scene, providing reliable predictions of result states under diverse actions. By rendering the images of the simulated outcomes, we enable the VLM to interface directly with the predicted future observations before they occur, which allows the VLM to bypass physical reasoning and instead focus on the semantic understanding and evaluation that it excels in.

To this end, we present \MethodName (\MethodAcronym), an approach that solves open-world manipulation in a model-predictive control (MPC) manner by integrating a VLM and an interactive digital twin. 
As shown in \Cref{fig:teaser}, given a video scan of the real-world environment, our method first builds an interactive digital twin with a controllable robot and interactable objects to enable physical simulation of possible outcomes of sampled actions.
In a sampling-based planning paradigm, our method generates candidate actions, simulates their outcomes within the digital twin, and utilizes the VLM to evaluate the resulting future states. 
Rather than requiring the VLM to reason about physics directly, we render RGB images from the digital twin to provide observations of predicted outcomes as visual prompts to the VLM. 
Unlike prior work that relies on fixed camera viewpoints, we adaptively adjust camera poses to optimize the visual inputs for the VLM, facilitating more effective spatial reasoning. 
We design a visual prompting mechanism that enables the VLM to assess the feasibility and desirability of different action outcomes, enabling planning of the optimal actions to solve the task.
Thereby, we enhance physical reasoning in VLM-driven robotic control without requiring additional robot data collection, VLM fine-tuning, or any in-context examples.

We evaluate \MethodAcronym in eight real-world manipulation tasks specified by natural language instructions. 
Our approach can effectively solve these tasks with success rates superior to baseline methods that use VLM for robotic control.
Additionally, we conduct thorough ablation studies and failure analysis to investigate the key factors contributing to our framework’s performance and robustness.

\section{Related Work}
\label{sec:related-work}
\noindent\textbf{VLM-based robotic control. }
Recent advances in VLMs have opened pathways toward open-world robot manipulation without any robotic demonstration by leveraging their strengths in semantic understanding and common-sense reasoning~\cite{driess2023palm, ahn2022can, pgvlm2024, wang2024vlm, patelreal}. 

However, VLMs, primarily trained on visual question-answering tasks, often struggle to reason about the physical effects of interactions necessary for motion planning. To address this limitation, prior works have explored diverse intermediate representations, such as reward functions, 2D keypoints, action vectors, and affordance~\cite{huang2023voxposer, fang2024moka, nasiriany2024pivot, huangrekep}, as outputs for VLMs. 
While these approaches show promising results, these settings diverge from VLMs' training distribution. Another line of work involves carefully designed chain-of-thought processes~\cite{duan2024manipulate, mu2024embodiedgpt, zawalski2024robotic}, which decompose tasks into smaller, predefined steps. Although effective for specific scenarios, this approach struggles to generalize to tasks that cannot be easily divided into such reasoning procedures and is limited by the inherent weaknesses of VLMs. 

In contrast, we aim to make open-world manipulation an in-distribution task for VLMs by leveraging digital twins to simulate future results. This allows VLMs to focus on evaluating possible future observations, aligning seamlessly with their strengths in description and evaluation.

\noindent\textbf{Model Predictive Control.}
Model Predictive Control (MPC) is a widely adopted optimization-based control strategy that has proven highly effective in robotics~\cite{finn2017deep, ebert2018visual, grandia2019feedback, nubert2020safe, minniti2021model}. MPC fundamentally consists of two key components: a dynamic model capable of predicting future system states given a sequence of actions, and an optimization process that determines the optimal action sequence by minimizing a predefined cost function. Recent MPC methods for open-world motion planning usually use images from a video generation model to predict the future results given different actions~\cite{zhao2024vlmpc, huang2023diffusion, zhao2024imaginenav}, leveraging the prior from the large-scale pre-training. However, video generation introduces artifacts and fails to generalize to complex scenarios that are far from the training distribution. Besides, generated video cannot guarantee to be precisely conditioned on the low-level actions. 

In this paper, we introduce physically grounded dynamic models by building digital twins of the real-world environments. Besides, compared with the traditional method of designing a cost function for each task, we employ VLMs for flexible and comprehensive evaluation over the results, enabling open-world manipulation on a wide range of tasks. 

\noindent\textbf{3D scene reconstruction for robot control.}
3D reconstruction methods, including neural radiance fields (NeRFs)~\cite{barron2021mip, barron2022mip, NeRF}, neural implicit surfaces~\cite{wang2021neus, wang2023neus2, oechsle2021unisurf}, and  Gaussian Splatting~\cite{huang20242d, kerbl20233d}, have revolutionized 3D scene modeling by enabling photorealistic rendering and dense geometry reconstruction. These methods provide a lot of potential for application in robot motion planning~\cite{wang2024d, shendistilled, kerrrobot, li20223d, qureshi2024splatsim}. Previous work leveraging scene reconstruction for motion planning either tries to distill 2D features generated by foundational models into 3D for spatial perception or enhance the demonstration to enable few-shot learning. However, these works usually assume the scene to be static and do not handle interactions over the reconstructed representation. 

Building digital twins from the real-world scenes, as a special branch of 3D scene reconstruction, enables a lot of applications in manipulation (i.e., real-to-sim-to-real method). However, previous methods usually regard the built digital twin as a virtual playground for collecting manipulation data~\cite{peng2024tiebot, wu2024rl, jiang2024dexmimicgen} or training robot policy through reinforcement learning~\cite{torne2024reconciling, patelreal, dai2024automated, torne2024robot, beltransliceit}, which are then transferred to the real world for manipulation.

In this work, we aim to enable dynamic and interactive modeling between the robot and its environment. To achieve this, we combine Gaussian splatting, known for its efficient and realistic rendering, with meshes for accurate physical modeling. Our hybrid representation leverages the strengths of both approaches, introducing interactive capabilities for robot control applications. This enables dynamic, physically grounded interactions in real-world scenarios, bridging the gap between static scene reconstruction and interactive manipulation. We leverage digital twin as a world model that can provide results of actions that have not happened yet in the real world. By combining with VLM as a critic, we enable a zero-shot setting for real-to-sim-to-real manipulation.

\section{Problem Formulation}
\label{sec:problem_formulation}
Our goal is to enable robots to perform manipulation tasks involving unseen objects and diverse goals.
We focus on complex scenarios that involve intricate contact, dynamics, and motion. These setups require robots to possess a thorough physical and semantic understanding of the environment.
We do not assume access to task-specific training data, in-context examples, or hard-coded motion primitives as used in prior work~\cite{huang2023voxposer, liang2023code, fang2024moka, kim2024openvla}. 
We consider a tabletop setting with one robotic arm. 
The framework's input consists of a natural language instruction $l$ specifying the task, and an RGB video scan $v$ of the scene. 
The output is an action sequence $\{a\}_{t=0}^{T-1}$ for achieving the task goal. 
Each action $a_t \in \mathbb{R}^7$ is defined as the 6-DoF gripper pose and the finger status (open or closed).

Central to our framework is a pre-trained vision–language model (VLM). The model processes an ordered sequence of interleaved text and RGB images and returns a textual response. We employ the VLM for various purposes with designed prompts.

\section{Method}
We propose to tackle open-world motion planning by complementing VLM's semantic reasoning ability with explicit modeling of dynamics through digital twins. Our framework builds two key components: 
1) We introduce a pipeline to automatically build interactive digital twins to support accurate modeling of diverse physical interactions and photorealistic rendering of the simulation outcomes. ~\Cref{sec:twin}. 
2) We formulate open-world manipulation as a model predictive control problem by prompting the VLM with futures provided by the digital twin, enabling adaptive observation and action optimization.~\Cref{sec:vlm}.

\begin{figure}[t]
    \centering
    \includegraphics[width=\linewidth]{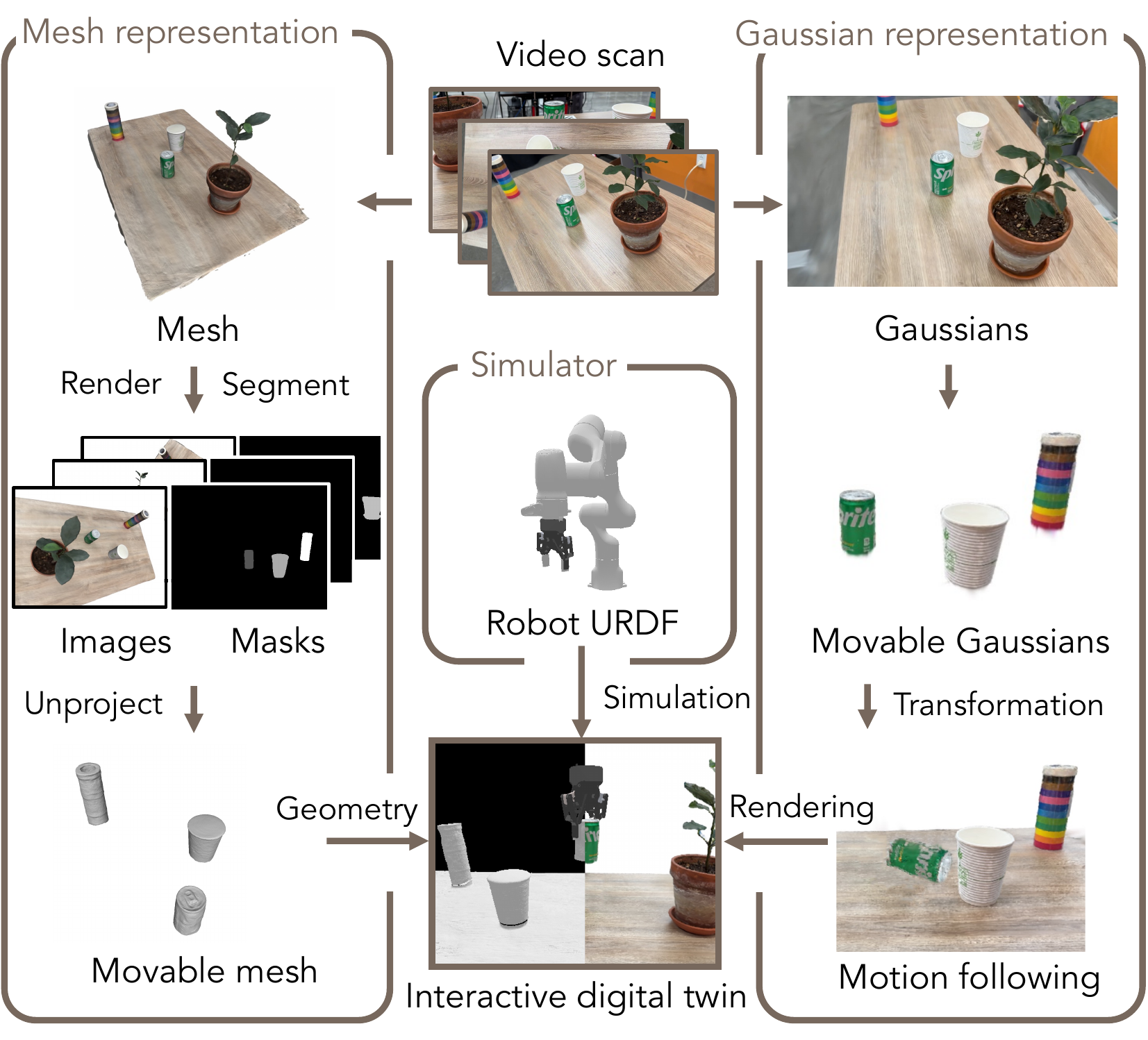}
    \caption{
    \textbf{Construction of interactive digital twins.} 
    Starting from a video scan of the environment, we construct an interactive digital twin that combines mesh-based simulation and Gaussian-based rendering.
The resulting twin enables photorealistic rendering and accurate simulation of object dynamics conditioned on robot actions.
    }
    \label{fig:twin}
\end{figure}

\subsection{Construction of Interactive Digital Twins}
\label{sec:twin}
To complement the semantic reasoning of the VLM with grounded physical predictions, we construct interactive digital twins that enable both accurate physical simulation and photorealistic rendering of future outcomes.
As shown in \Cref{fig:twin}, our construction pipeline consists of two key stages: (1) reconstructing scenes with accurate geometry and visual appearance, and (2) making the scene interactable to support robot actions and environment dynamics.

Unlike prior work, which often focuses solely on static reconstruction~\cite{shen2023distilled, kerrrobot}, our method produces dynamic, action-conditioned digital twins by combining mesh-based physical modeling with efficient Gaussian splatting for rendering. This hybrid design supports fine-grained simulation of physical interactions while maintaining high-quality visual fidelity.

\begin{figure*}[t]
    \centering
    \includegraphics[width=\linewidth]{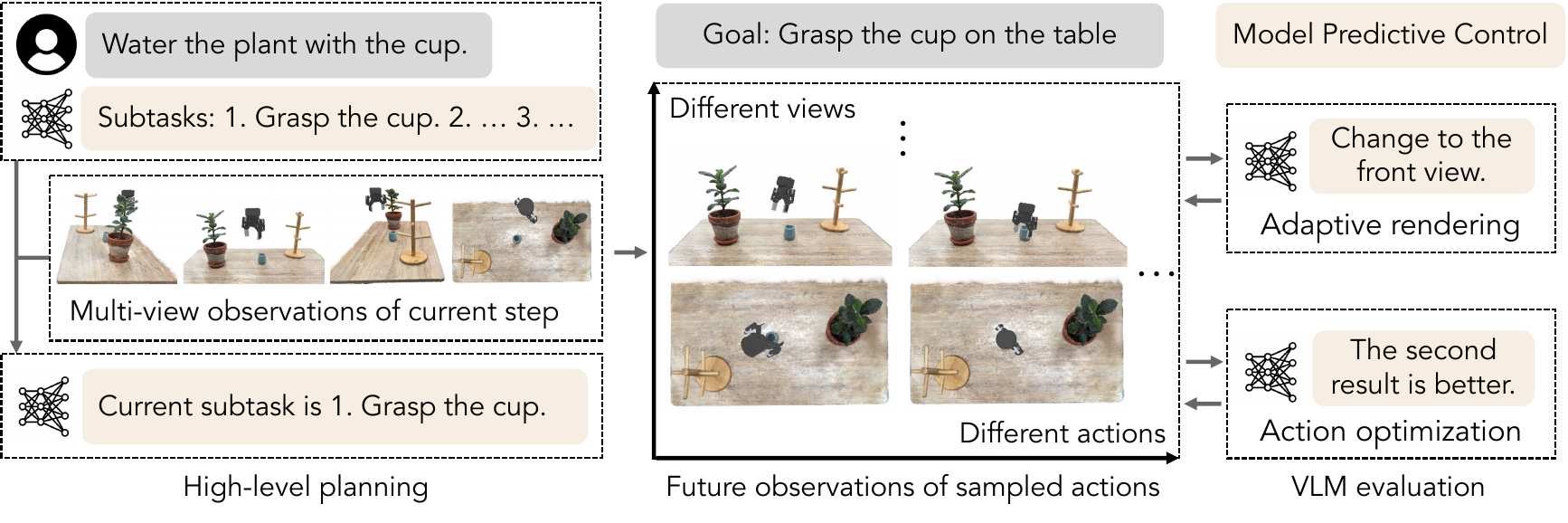}
    \caption{
    \textbf{Model Predictive Control through Simulation-Informed Prompting.} Given a free-form instruction, our framework first performs high-level planning by generating structured subtasks from multi-view observations.
At each step, the interactive digital twin simulates future states of candidate actions and renders the outcomes.
The VLM adaptively selects the most informative view for rendering and evaluates the predicted outcomes for sampling-based motion planning.
}
    \label{fig:mpc}
\end{figure*}

\noindent \textbf{Hybrid reconstruction: } 
Given a video scan of the real-world environment, we first reconstruct a hybrid scene representation that achieves both geometric accuracy and photorealistic appearance (\Cref{fig:twin}).
We apply 2D Gaussian Splatting~\cite{huang20242d} to create a Gaussian representation $G$ supervised by the extracted RGB frames. To recover precise scene geometry, we convert $G$ into a mesh representation $M$ through TSDF volume integration, preserving fine details necessary for physical simulation.

\noindent \textbf{Interactive objects:} 
To model dynamics for robotic manipulation, the reconstructed scene must model movable objects.
Starting from $M$, we render multi-view images of the scene and use Molmo~\cite{molmo2024}, a vision-language model (VLM), to identify key movable objects based on task instructions $l$. Given Molmo’s 2D keypoints, we employ SAM2~\cite{ravi2024sam2} to track segmentation masks across views. These 2D masks are then projected back to 3D, labeling mesh vertices and segmenting out movable object meshes within $M$.

\noindent \textbf{Realistic rendering:}  
In addition to movable meshes for geometric information, we also need high visual fidelity and ensure realistic rendering in accordance with object movement. 
To achieve realistic and responsive rendering, we enable the Gaussians $G$ to dynamically follow the movement of their associated object meshes.
Specifically, we first associate each Gaussian point with the nearest movable mesh based on the Euclidean distance between their centers. During simulation, the anchored Gaussians inherit the translation and rotation of their corresponding meshes, ensuring that the rendered appearance remains consistent with object motions.

\noindent \textbf{Physical simulation:} 
Finally, we integrate a physics simulator $S$~\cite{gu2023maniskill2} equipped with the robot’s URDF $U$ to model dynamics under interaction. The simulator computes physically plausible state transitions when applying diverse candidate actions, ensuring the digital twin can predict action-conditioned outcomes with high fidelity.

Through this construction pipeline, we obtain an interactive digital twin where the mesh representation provides physical structure, the Gaussian splatting enables efficient and realistic rendering, and the simulator governs the dynamics. Additional implementation details are provided in the supplementary material.

\subsection{Simulation and Rendering for Visual Prompting}
\label{sec:prediction_of_future_observations}

We simulate and render future outcomes in the digital twin to generate visual prompts for the vision-language model (VLM).
Given a robot action $a \in \mathbb{R}^7$, the simulator $S$ predicts the resulting transformation of the robot and scene objects within the mesh representation $M$, denoted as $\mathcal{T} = S(M, a)$.
We apply this transformation $\mathcal{T}$ to the Gaussian scene representation $G$, obtaining an updated state $G(\mathcal{T})$ that encodes the physical consequences of the action.

While VLMs operate purely on 2D images and inherently lack strong spatial reasoning, we facilitate 3D understanding by synthesizing multi-view observations of each predicted future.
Given the updated state $G(\mathcal{T})$, we render background scene images from a set of camera configurations $C$, producing $I_g = \text{Render}(G(\mathcal{T}), C)$, which depict the environment without the robot.
Separately, we render the robot executing action $a$ using the simulator and its URDF model, yielding $I_r = \text{Render}(S(U, a))$.
The background and robot images are composited based on depth information to generate the final multi-view observations $I = { I_i \in \mathbb{R}^{W \times H \times 3} }$.
These synthesized observations allow the VLM to perceive the manipulation scene from diverse viewpoints, enhancing its ability to reason about future states and improving the accuracy of action selection.

\subsection{Motion Planning via Simulation-Informed Prompting}
\label{sec:vlm}
We formulate the control problem as a model predictive control (MPC) framework, where the vision-language model (VLM) serves exclusively as an evaluator of predicted future states.
Our approach decomposes tasks into subgoals, adaptively selects viewpoints to facilitate VLM reasoning, and optimizes actions through sampling-based planning.

{
\setlength{\tabcolsep}{7.6mm}{
\begin{table*}[t]
    \centering
    \begin{tabular}{lll}  
    \toprule

    Tasks & Instruction & Success Criteria  \\
    \midrule
    Water plant   &  Water the plant with the cup  & The cup is over the plant pot with tilting action \\
    Play drum  & Hit the drum with the drum stick & The head of the drum stick contacts the drum \\
    Clean up & Wipe the tea with the sponge  & The sponge makes contact with the spilled tea \\
    Press spacebar   &  Press the space on the keyboard  & The spacebar is pressed \\
    Cucumber basket    &  Put the green cucumber into the basket  & The cucumber is in the basket \\
    Pair up shoes   &   Pair up the shoes & Corresponding shoes are next to each other in parallel\\
    Unplug charger     &  Unplug the charger  & The charger is no longer plugged into the power strip \\
    Lowest tune     &  Play the lowest pitch with the drum stick  & The head of the drum stick contacts the lowest key \\
    \bottomrule
    \end{tabular}
    \caption{\textbf{Definition of tasks}. We list the text instructions for prompting VLM and the success criteria of each task.}
    \label{tab:tasks}
\end{table*}
}
}

\noindent\textbf{Subgoal Decomposition.}
We first leverage the semantic reasoning capabilities of the VLM to decompose complex tasks into structured subgoals.
Given initial multi-view observations $I_0$ and a task instruction $l$, the VLM generates a set of subtasks $\tau = {\tau_1, \tau_2, \dots}$, each specifying an intermediate objective.
At each planning step $t$, the VLM is prompted with the current observations $I_t$ and the subtask set $\tau$, and selects the active subgoal $\tau_i$ corresponding to the agent's current progress.
This decomposition localizes the optimization objective, improving sample efficiency and enhancing planning robustness.

\noindent\textbf{Adaptive Viewpoint Selection.}
Since robot actions are defined in $SE(3)$ space but the VLM operates solely on 2D visual inputs, spatial reasoning critically depends on viewpoint selection.
At each planning step, we render observations of the updated scene $G(\mathcal{T})$ under a set of candidate camera configurations $C$.
Conditioned on $\tau_i$ and $I_t$, the VLM selects the viewpoint $C_t$ that provides the most informative observation for distinguishing between action outcomes.
This adaptive rendering procedure strengthens the VLM’s ability to reason about geometric and physical variations essential for task success (see Figure~\ref{fig:mpc}).

\noindent\textbf{Sampling-Based Planning.}
With the active subgoal $\tau_i$ and selected viewpoint $C_t$ determined, the framework proceeds to low-level action generation.
We employ the Cross-Entropy Method (CEM)~\cite{rubinstein1999cross} for structured action optimization, by leveraging the digital twins to explicitly model the dynamics and the VLM to evaluate the predicted outcomes.
To select the action, candidate actions ${a_k}$ are initially sampled from a multivariate Gaussian distribution and simulated within the digital twin to obtain their corresponding future observations $o_{t,k}$.
The simulated outcomes ${o_{t,k}}$ are partitioned into $m$ groups, each individually prompted to the VLM along with the subgoal $\tau_i$, and the VLM selects the outcome that best advances the subgoal.
The corresponding elite actions are then used to update the sampling distribution.
This sampling and refinement process is repeated for three iterations, after which the mean of the final distribution is taken as the optimized action $a_t$.

\begin{table*}[t]
    \centering
    \begin{tabular}{lcccccccc}  
    \toprule

    Methods & Water plant & Play drum & Press spacebar & Pair up shoes & Cucumber basket & Lowest tune & Unplug charger & Clean up \\
    \midrule
    Voxposer\footnotemark[1]~\cite{huang2023voxposer}   &     \textbf{6/10}       & 2/10  & 0/10 & 2/10 & 8/10 & 0/10 & 4/10 & 0/10 \\
    MOKA~\cite{fang2024moka}      & 4/10 & 5/10 & 0/10 & 2/10 & 5/10 & 0/10 & 0/10 & 5/10 \\
    OpenVLA~\cite{kim2024openvla} &    1/10 & 0/10 & 1/10 & 0/10 & 4/10 & 0/10 & 0/10 & 0/10 \\
    $\pi_0$~\cite{black2024pi_0}  & 0/10 & 0/10 & 0/10 & 1/10 & 3/10  & 0/10 & 2/10 & 0/10 \\
    OpenVLA-finetuned & 0/10 & 0/10 & 0/10 & 0/10 & 5/10 & 0/10 & 0/10 & 0/10 \\
    $\pi_0$-finetuned      & 0/10 & 0/10 & 0/10 & 2/10 & 2/10 & 0/10 & 3/10 & 2/10 \\
    Ours     &  5/10 & \textbf{6/10}  & \textbf{5/10} & \textbf{6/10} & \textbf{9/10} & \textbf{4/10} & \textbf{7/10} & \textbf{5/10} \\
    \bottomrule
    \end{tabular}
    \caption{\textbf{Comparison against baselines}. \MethodAcronym better leverages the reasoning ability of VLM and improves the performance on most of the tasks. Besides, our image-based evaluation on results provides more flexibility than value map or key points, enabling challenging tasks that are hard to be parametrized by previous representations (e.g., play the lowest tune).}
    \label{tab:experiment}
    \vspace{2mm}
\end{table*}

\vspace{2mm}
\section{Experiments} 
\label{sec:experiments}
\vspace{1mm}

To validate the effectiveness of our framework, in this section, we design eight real-world manipulation tasks that require 6 DoF control, semantic understanding, and diverse manipulation skills. 
We compare our approach against prior works on open-world manipulation that leverage VLMs or train on large-scale robotic data. Additionally, we conduct ablation experiments to evaluate the contribution of each component to the overall performance.

\vspace{2mm}
\subsection{Experimental setup}
\vspace{2mm}
\noindent\textbf{Tasks.}
As shown in Table~\ref{tab:tasks}, we introduce eight manipulation tasks that require intricate understanding of the physical world and diverse manipulation skills: \texttt{water plant}, \texttt{play drum}, \texttt{press spacebar}, \texttt{pair up shoes}, \texttt{cucumber basket}, \texttt{lowest tune}, \texttt{unplug charger}, and \texttt{clean up}.
For each task, we construct five manipulation scenes, featuring randomized object layouts and different distractors. Please see the supplementary material for more details about task design.

\vspace{1mm}
\noindent\textbf{Baselines.} 
We compare our model against several state-of-the-art methods. 
\textbf{VoxPoser}*~\cite{huang2023voxposer} leverages a VLM to predict 3D value map for motion optimization. 
We enhance it by providing ground-truth segmented object point clouds from our digital twin, significantly improving its perception accuracy.
\textbf{MOKA}~\cite{fang2024moka} chooses the 2D keypoints as intermediate representations for VLM to predict, which are then converted into actions based on the depth information from a depth camera. \textbf{OpenVLA}~\cite{kim2024openvla}  is a 7B-parameter open-source vision-language-action model fine-tuned from a VLM using 970k real-world robot demonstrations~\cite{o2023open}. $\bm{\pi_0}$~\cite{black2024pi_0}, a state-of-the-art vision-language-action model trained on diverse robot demonstrations. For both OpenVLA and $\bm{\pi_0}$, we report their performances under a zero-shot setting and after task-specific fine-tuning on 20 expert demonstrations for each task.

\vspace{1mm}
\noindent\textbf{Metrics.} We use the success rate as the evaluation metric. A task is considered a failure if the robot causes irreversible results or if the maximum step budget or time limit is reached. The task is successful if the success criteria are met. Please see the supplementary material for more details.

\vspace{1mm}
\noindent\textbf{Implementation details.} 
We adopt GPT-4o~\cite{achiam2023gpt} for both our method and the baselines. During inference, we render four camera views and allow VLM to select observations from these perspectives. For the CEM optimization, we use 3 iterations with 90 samples per iteration. 
The planning policies are rolled out twice per scene to consider the randomness in VLM planning, resulting in 10 trials per task in total. Please see the supplementary material for more details about task and baseline design.


\vspace{2mm}
\subsection{Quantitative results}
\vspace{2mm}
\Cref{tab:experiment} compares the success rates of our method against those of the baselines. 
Since Voxposer and MOKA rely on open-vocabulary detectors to detect objects before manipulation, they fail when the perception system cannot recognize specific object parts, such as the spacebar on a keyboard or the lowest key on a xylophone. In contrast, our method directly leverages the VLM to comprehend and reason about simulated future states, eliminating the need for a perception module and resulting in greater robustness.
As for OpenVLA and $\pi_0$, while they can perform zero-shot on simple tasks due to their training on large-scale robotic datasets, their generalization is limited by the coverage of the training data, making them less effective for complex tasks. Even with task-specific demonstrations, the trained methods still struggle to generalize to unseen layouts. We hypothesize that more data is needed for trained methods. In contrast, our method benefits from the commonsense reasoning capabilities of the VLM, enabling broader task coverage and adaptability. We particularly excel in tasks requiring precise gripper pose alignment, as our approach allows for simulation-based ``rehearsal'' before execution.

\begin{figure}[t]
    \centering
    \includegraphics[width=\linewidth]{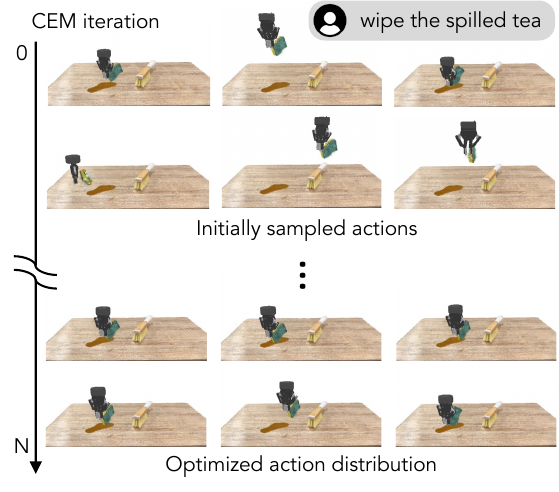}
    \vspace{-5mm}
    \caption{\textbf{Example on action optimization.} We show the action optimization results of one planning step in subtask ``wipe the spilled tea". Our digital twin could simulate diverse results with accurate motion and collision of the sponge in initial sampling and VLM could effectively optimize the action distribution to move the sponge towards the tea.}
    \label{fig:optimize}
\end{figure}

\begin{figure*}[!ht]
    \centering
    \includegraphics[width=\linewidth]{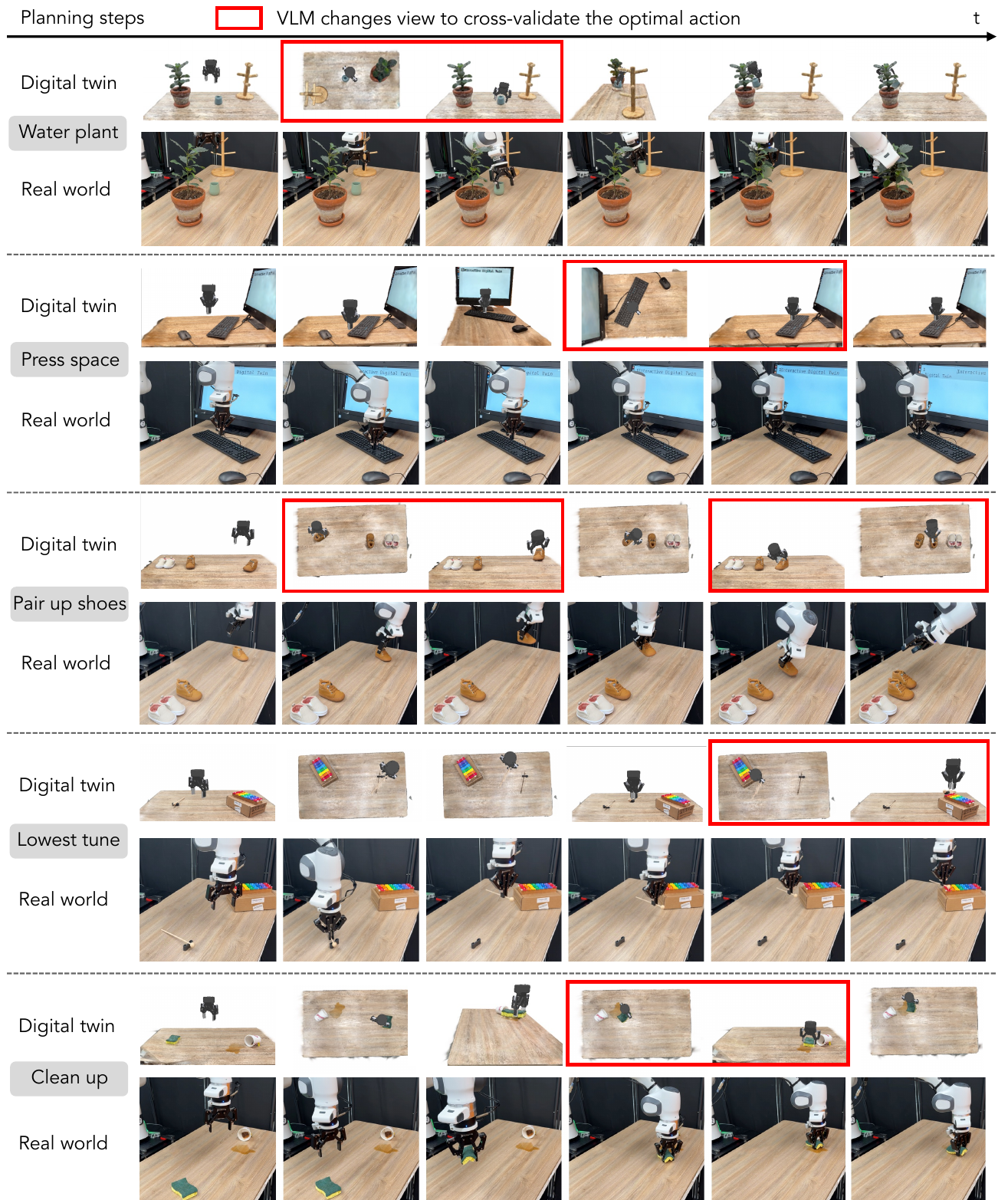}
    \caption{\textbf{Example trajectories.} Example trajectories planned by our framework in both digital twin and real world (aligned). We highlight some key steps where VLM chooses to adaptively change the rendering view for better result evaluation (e.g., aligning the gripper from different perspectives for grasping, pressing, placing or hitting).}
    \label{fig:results}
    \vspace{-2mm}
\end{figure*}

\vspace{2mm}
\subsection{Qualitative results}
\vspace{2mm}
We visualize the action optimization process for a single planning step in the ``clean up'' task in \Cref{fig:optimize}. Initially, the digital twin simulates a diverse set of actions with precise dynamics (\emph{e.g.}, object rotating due to grasping, object dropping due to collisions, etc.) and photorealistic rendering. Based on the simulation results and the task instruction, the VLM selects the results that better align with the target objective. An action distribution is fit to the selected elite actions, from which new actions are resampled for further simulation in the digital twin. This iterative process leads to an optimized action distribution that aligns more closely with the goal of wiping the spilled tea using the sponge.

\begin{table*}[ht]
\vspace{7mm}
    \centering
    \begin{tabular}{lcccccccc}  
    \toprule

    Methods & Water plant & Play drum  & Press spacebar & Pair up shoes & Cucumber basket & Lowest tune & Unplug charger &  Clean up \\
    \midrule
    w/o Views         &  1/10 & 0/10  & 0/10 & 3/10 & 8/10 & 0/10 & 3/10 & 1/10 \\
    w/o Subtasks      &  5/10 & 0/10  & \textbf{6/10} & 3/10 & 8/10 & 2/10 & 7/10  & 0/10 \\
    w/o CEM & 0/10 & 2/10  & 2/10 & 2/10 & 1/10 & 0/10  & 4/10 & 0/10 \\
    
    Ours Full Method    &    \textbf{5/10}     & \textbf{6/10}  & 5/10 & \textbf{6/10} & \textbf{9/10} & \textbf{4/10} & \textbf{7/10} & \textbf{5/10} \\
    \bottomrule
    \end{tabular}
    \caption{\textbf{Ablation study.} We validate the effectiveness of our components. Multi-view observations are essential for tasks that are sensitive to perspectives. Subtask division improves the performance on tasks that require multi-stage planning. Most importantly, CEM significantly increases the sampling efficiency, facilitates effective planning.}
    \label{tab:ablations}
    \vspace{3mm}
\end{table*}

\Cref{fig:results} shows some examples of the rollout trajectories planned by our framework in both digital twins and the real world. By mirroring possible interactions in the simulated world, our framework provides a flexible and effective way for VLMs to guide the motion of the robot on diverse tasks in an open-world environment. The digital twin and the real world are aligned based on planning steps. We highlight key planning steps where VLM chooses to change the observation view to better assess the results, showing the benefits of our adaptive rendering design. Please see the supplementary material for more visualization and videos.

\vspace{2mm}
\subsection{Ablation study}
\vspace{2mm}
To assess the contribution of each component in our framework, we begin with the full system and systematically remove each component in turn.
In the ``w/o views'' setting, we fix the camera to a static top-down perspective instead of allowing the VLM to select a view for each planning step.
For ``w/o subtasks,'' we use the user instruction directly as the goal for each planning step rather than first generating subtasks.
In the ``w/o CEM" setting, we simply take the mean value of the selected actions without optimizing the action distribution or resampling.

As shown in \Cref{tab:ablations}, while performance varies across different tasks due to their diverse requirements, our full method achieves the best results in most of the tasks. Multi-view observations, enabled by our adaptive rendering, enhance spatial perception, particularly in tasks requiring precise control (e.g., Press spacebar, Play the drum). Dividing tasks into subtasks allows the VLM to focus on more concrete evaluation criteria for each state, which is beneficial for tasks involving multiple stages, such as playing the drum and cleaning up. Lastly, the CEM process significantly improves sampling efficiency, producing action distributions that better align with the goal, which in general contributes the most to our model predictive control framework.

\begin{figure}[t]
    \centering
    \includegraphics[width=\linewidth]{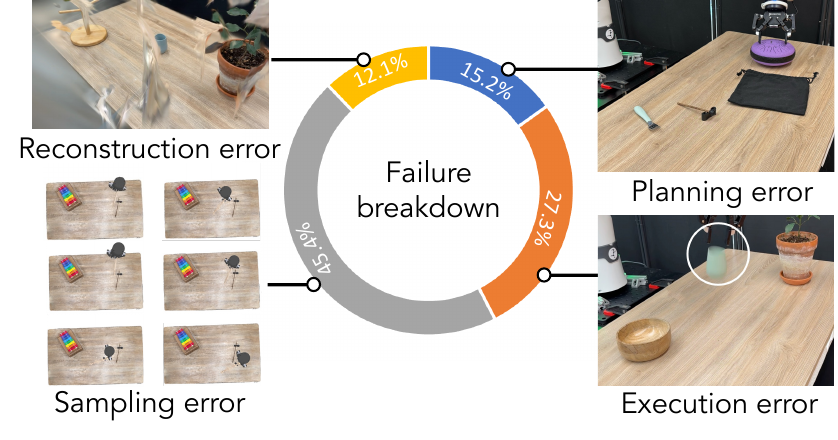}
    \caption{\textbf{Failure analysis.} Our main failure cases can be divided into four categories. We show the percentage and provide an example for each failure type.}
    \label{fig:failure}
\end{figure}

\vspace{2mm}
\subsection{Failure breakdown}
\vspace{2mm}
The failure cases can be categorized into four groups:  
\begin{itemize}
    \item \textbf{Reconstruction error:} The quality of our digital twin depends on the accuracy of camera pose estimation and 3D reconstruction. Noisy in poses or optimization often result in visual artifacts, which can reduce the effectiveness of the VLM.
    \item \textbf{Planning error:} When subtasks are not properly defined or the model fails to recognize the current stage, the robot may execute actions incorrectly. For example, the robot may attempt to move the gripper directly to the drum without first picking up the drumstick.
    \item \textbf{Execution error:} In the real world, the gripper may misalign during grasping or drop objects during movement due to insufficient grip or friction.
    \item \textbf{Sampling error:} This is the primary source of failure in our framework. Due to the inherent randomness of action sampling and errors in VLM reasoning, the system may struggle to sample actions to achieve the goal within a limited step budget.
\end{itemize}
We show their percentage and some examples in \Cref{fig:failure}. Please see
the supplementary material for a more detailed failure breakdown.

\vspace{2mm}
\section{Limitations} 
\vspace{1mm}
\label{sec:limitation}
This work explores the potential of using vision-language models (VLMs) for open-world robotic manipulation without relying on robotic data or in-context examples. While the approach shows promise, several limitations remain.

A primary limitation lies in computational efficiency. Reconstructing 3D representations from real-world scenes takes about 20 minutes per scene. Inference through prompting VLMs also introduces latency due to token generation. These factors limit the system’s applicability to real-time settings. Progress in 3D reconstruction methods and the use of smaller, open-source VLMs may help alleviate these issues. Please see the supplementary material about the discussion and improvement on efficiency.

Second, the performance is bounded by the capability of current VLMs. Inaccuracies in perception can lead to suboptimal evaluations of candidate actions, introducing variance into planning. Nevertheless, the system is designed to benefit directly from advances in VLMs, and we expect robustness and efficiency to improve as models evolve.
\vspace{2mm}
\section{Conclusion} 
\vspace{1mm}
\label{sec:conclusion}
We propose a framework that combines vision-language models (VLMs) with interactive digital twins for open-world robotic manipulation. The system builds hybrid representations from real-world scans and enables physical interactions, supporting explicit modeling of dynamics and photorealistic rendering for action selection with VLMs. By using MPC with adaptive rendering, the method leverages VLMs' high-level semantic understanding while grounding low-level action optimization in physical simulation. Experiments across a range of tasks show that the framework outperforms prior VLM- and vision-language-action model (VLA-based) approaches, demonstrating a viable path toward closing the gap between semantic reasoning and robotic control.
\section{Acknowledgement} 
\label{sec:acknowledgement}
The research is partially supported by a gift from Ai2, a NVIDIA Academic Grant, and DARPA TIAMAT program No. HR00112490422. Its contents are solely the responsibility of the authors and do not necessarily represent the official views of DARPA.


\bibliographystyle{plainnat}
\bibliography{main}

\begin{thebibliography}{55}
\providecommand{\natexlab}[1]{#1}
\providecommand{\url}[1]{\texttt{#1}}
\expandafter\ifx\csname urlstyle\endcsname\relax
  \providecommand{\doi}[1]{doi: #1}\else
  \providecommand{\doi}{doi: \begingroup \urlstyle{rm}\Url}\fi

\bibitem[Achiam et~al.(2023)Achiam, Adler, Agarwal, Ahmad, Akkaya, Aleman, Almeida, Altenschmidt, Altman, Anadkat, et~al.]{achiam2023gpt}
Josh Achiam, Steven Adler, Sandhini Agarwal, Lama Ahmad, Ilge Akkaya, Florencia~Leoni Aleman, Diogo Almeida, Janko Altenschmidt, Sam Altman, Shyamal Anadkat, et~al.
\newblock Gpt-4 technical report.
\newblock \emph{arXiv:2303.08774}, 2023.

\bibitem[Ahn et~al.(2022)Ahn, Brohan, Brown, Chebotar, Cortes, David, Finn, Fu, Gopalakrishnan, Hausman, et~al.]{ahn2022can}
Michael Ahn, Anthony Brohan, Noah Brown, Yevgen Chebotar, Omar Cortes, Byron David, Chelsea Finn, Chuyuan Fu, Keerthana Gopalakrishnan, Karol Hausman, et~al.
\newblock Do as i can, not as i say: Grounding language in robotic affordances.
\newblock \emph{arXiv:2204.01691}, 2022.

\bibitem[Barron et~al.(2021)Barron, Mildenhall, Tancik, Hedman, Martin-Brualla, and Srinivasan]{barron2021mip}
Jonathan~T Barron, Ben Mildenhall, Matthew Tancik, Peter Hedman, Ricardo Martin-Brualla, and Pratul~P Srinivasan.
\newblock Mip-nerf: A multiscale representation for anti-aliasing neural radiance fields.
\newblock In \emph{ICCV}, 2021.

\bibitem[Barron et~al.(2022)Barron, Mildenhall, Verbin, Srinivasan, and Hedman]{barron2022mip}
Jonathan~T Barron, Ben Mildenhall, Dor Verbin, Pratul~P Srinivasan, and Peter Hedman.
\newblock Mip-nerf 360: Unbounded anti-aliased neural radiance fields.
\newblock In \emph{CVPR}, 2022.

\bibitem[Beltran-Hernandez et~al.(2024)Beltran-Hernandez, Erbetti, and Hamaya]{beltransliceit}
Cristian~Camilo Beltran-Hernandez, Nicolas Erbetti, and Masashi Hamaya.
\newblock Sliceit!: Simulation-based reinforcement learning for compliant robotic food slicing.
\newblock In \emph{ICRA Workshop}, 2024.

\bibitem[Black et~al.(2024)]{black2024pi_0}
Kevin Black et~al.
\newblock $\pi_0$: A vision-language-action flow model for general robot control.
\newblock \emph{arXiv:2410.24164}, 2024.

\bibitem[Brohan et~al.(2023)Brohan, Brown, Carbajal, Chebotar, Chen, Choromanski, Ding, Driess, Dubey, Finn, et~al.]{brohan2023rt}
Anthony Brohan, Noah Brown, Justice Carbajal, Yevgen Chebotar, Xi~Chen, Krzysztof Choromanski, Tianli Ding, Danny Driess, Avinava Dubey, Chelsea Finn, et~al.
\newblock Rt-2: Vision-language-action models transfer web knowledge to robotic control.
\newblock \emph{arXiv:2307.15818}, 2023.

\bibitem[Dai et~al.(2024)Dai, Wong, Jiang, Wang, Gokmen, Zhang, Wu, and Fei-Fei]{dai2024automated}
Tianyuan Dai, Josiah Wong, Yunfan Jiang, Chen Wang, Cem Gokmen, Ruohan Zhang, Jiajun Wu, and Li~Fei-Fei.
\newblock Automated creation of digital cousins for robust policy learning.
\newblock \emph{arXiv:2410.07408}, 2024.

\bibitem[Deitke et~al.(2024{\natexlab{a}})Deitke, Clark, Lee, Tripathi, Yang, Park, Salehi, Muennighoff, Lo, Soldaini, Lu, Anderson, Bransom, Ehsani, Ngo, Chen, Patel, Yatskar, Callison-Burch, Head, Hendrix, Bastani, VanderBilt, Lambert, Chou, Chheda, Sparks, Skjonsberg, Schmitz, Sarnat, Bischoff, Walsh, Newell, Wolters, Gupta, Zeng, Borchardt, Groeneveld, Dumas, Nam, Lebrecht, Wittlif, Schoenick, Michel, Krishna, Weihs, Smith, Hajishirzi, Girshick, Farhadi, and Kembhavi]{molmo2024}
Matt Deitke, Christopher Clark, Sangho Lee, Rohun Tripathi, Yue Yang, Jae~Sung Park, Mohammadreza Salehi, Niklas Muennighoff, Kyle Lo, Luca Soldaini, Jiasen Lu, Taira Anderson, Erin Bransom, Kiana Ehsani, Huong Ngo, YenSung Chen, Ajay Patel, Mark Yatskar, Chris Callison-Burch, Andrew Head, Rose Hendrix, Favyen Bastani, Eli VanderBilt, Nathan Lambert, Yvonne Chou, Arnavi Chheda, Jenna Sparks, Sam Skjonsberg, Michael Schmitz, Aaron Sarnat, Byron Bischoff, Pete Walsh, Chris Newell, Piper Wolters, Tanmay Gupta, Kuo-Hao Zeng, Jon Borchardt, Dirk Groeneveld, Jen Dumas, Crystal Nam, Sophie Lebrecht, Caitlin Wittlif, Carissa Schoenick, Oscar Michel, Ranjay Krishna, Luca Weihs, Noah~A. Smith, Hannaneh Hajishirzi, Ross Girshick, Ali Farhadi, and Aniruddha Kembhavi.
\newblock Molmo and pixmo: Open weights and open data for state-of-the-art multimodal models.
\newblock \emph{arXiv:2409.17146}, 2024{\natexlab{a}}.

\bibitem[Deitke et~al.(2024{\natexlab{b}})Deitke, Clark, Lee, Tripathi, Yang, Park, Salehi, Muennighoff, Lo, Soldaini, et~al.]{deitke2024molmo}
Matt Deitke, Christopher Clark, Sangho Lee, Rohun Tripathi, Yue Yang, Jae~Sung Park, Mohammadreza Salehi, Niklas Muennighoff, Kyle Lo, Luca Soldaini, et~al.
\newblock Molmo and pixmo: Open weights and open data for state-of-the-art multimodal models.
\newblock \emph{arXiv:2409.17146}, 2024{\natexlab{b}}.

\bibitem[Driess et~al.(2023)Driess, Xia, Sajjadi, Lynch, Chowdhery, Ichter, Wahid, Tompson, Vuong, Yu, et~al.]{driess2023palm}
Danny Driess, Fei Xia, Mehdi~SM Sajjadi, Corey Lynch, Aakanksha Chowdhery, Brian Ichter, Ayzaan Wahid, Jonathan Tompson, Quan Vuong, Tianhe Yu, et~al.
\newblock Palm-e: An embodied multimodal language model.
\newblock \emph{arXiv:2303.03378}, 2023.

\bibitem[Duan et~al.(2024)Duan, Yuan, Pumacay, Wang, Ehsani, Fox, and Krishna]{duan2024manipulate}
Jiafei Duan, Wentao Yuan, Wilbert Pumacay, Yi~Ru Wang, Kiana Ehsani, Dieter Fox, and Ranjay Krishna.
\newblock Manipulate-anything: Automating real-world robots using vision-language models.
\newblock \emph{arXiv:2406.18915}, 2024.

\bibitem[Ebert et~al.(2018)Ebert, Finn, Dasari, Xie, Lee, and Levine]{ebert2018visual}
Frederik Ebert, Chelsea Finn, Sudeep Dasari, Annie Xie, Alex Lee, and Sergey Levine.
\newblock Visual foresight: Model-based deep reinforcement learning for vision-based robotic control.
\newblock \emph{arXiv:1812.00568}, 2018.

\bibitem[Fang et~al.(2024)Fang, Liu, Abbeel, and Levine]{fang2024moka}
Kuan Fang, Fangchen Liu, Pieter Abbeel, and Sergey Levine.
\newblock Moka: Open-vocabulary robotic manipulation through mark-based visual prompting.
\newblock In \emph{RSS}, 2024.

\bibitem[Finn and Levine(2017)]{finn2017deep}
Chelsea Finn and Sergey Levine.
\newblock Deep visual foresight for planning robot motion.
\newblock In \emph{ICRA}, 2017.

\bibitem[Gao et~al.(2024)Gao, Sarkar, Xia, Xiao, Wu, Ichter, Majumdar, and Sadigh]{pgvlm2024}
Jensen Gao, Bidipta Sarkar, Fei Xia, Ted Xiao, Jiajun Wu, Brian Ichter, Anirudha Majumdar, and Dorsa Sadigh.
\newblock Physically grounded vision-language models for robotic manipulation.
\newblock In \emph{ICRA}, 2024.

\bibitem[Grandia et~al.(2019)Grandia, Farshidian, Ranftl, and Hutter]{grandia2019feedback}
Ruben Grandia, Farbod Farshidian, Ren{\'e} Ranftl, and Marco Hutter.
\newblock Feedback mpc for torque-controlled legged robots.
\newblock In \emph{IROS}, 2019.

\bibitem[Gu et~al.(2023)Gu, Xiang, Li, Ling, Liu, Mu, Tang, Tao, Wei, Yao, Yuan, Xie, Huang, Chen, and Su]{gu2023maniskill2}
Jiayuan Gu, Fanbo Xiang, Xuanlin Li, Zhan Ling, Xiqiang Liu, Tongzhou Mu, Yihe Tang, Stone Tao, Xinyue Wei, Yunchao Yao, Xiaodi Yuan, Pengwei Xie, Zhiao Huang, Rui Chen, and Hao Su.
\newblock Maniskill2: A unified benchmark for generalizable manipulation skills.
\newblock In \emph{ICLR}, 2023.

\bibitem[Huang et~al.(2024{\natexlab{a}})Huang, Yu, Chen, Geiger, and Gao]{huang20242d}
Binbin Huang, Zehao Yu, Anpei Chen, Andreas Geiger, and Shenghua Gao.
\newblock 2d gaussian splatting for geometrically accurate radiance fields.
\newblock In \emph{SIGGRAPH}, 2024{\natexlab{a}}.

\bibitem[Huang et~al.(2023{\natexlab{a}})Huang, Wang, Li, Jia, Liu, Zhu, Liang, and Zhu]{huang2023diffusion}
Siyuan Huang, Zan Wang, Puhao Li, Baoxiong Jia, Tengyu Liu, Yixin Zhu, Wei Liang, and Song-Chun Zhu.
\newblock Diffusion-based generation, optimization, and planning in 3d scenes.
\newblock In \emph{CVPR}, 2023{\natexlab{a}}.

\bibitem[Huang et~al.(2023{\natexlab{b}})Huang, Wang, Zhang, Li, Wu, and Fei-Fei]{huang2023voxposer}
Wenlong Huang, Chen Wang, Ruohan Zhang, Yunzhu Li, Jiajun Wu, and Li~Fei-Fei.
\newblock Voxposer: Composable 3d value maps for robotic manipulation with language models.
\newblock In \emph{CoRL}, 2023{\natexlab{b}}.

\bibitem[Huang et~al.(2024{\natexlab{b}})Huang, Wang, Li, Zhang, and Fei-Fei]{huangrekep}
Wenlong Huang, Chen Wang, Yunzhu Li, Ruohan Zhang, and Li~Fei-Fei.
\newblock Rekep: Spatio-temporal reasoning of relational keypoint constraints for robotic manipulation.
\newblock In \emph{CoRL}, 2024{\natexlab{b}}.

\bibitem[Jiang et~al.(2024)Jiang, Xie, Lin, Xu, Wan, Mandlekar, Fan, and Zhu]{jiang2024dexmimicgen}
Zhenyu Jiang, Yuqi Xie, Kevin Lin, Zhenjia Xu, Weikang Wan, Ajay Mandlekar, Linxi Fan, and Yuke Zhu.
\newblock Dexmimicgen: Automated data generation for bimanual dexterous manipulation via imitation learning.
\newblock \emph{arXiv:2410.24185}, 2024.

\bibitem[Kerbl et~al.(2023)Kerbl, Kopanas, Leimk{\"u}hler, and Drettakis]{kerbl20233d}
Bernhard Kerbl, Georgios Kopanas, Thomas Leimk{\"u}hler, and George Drettakis.
\newblock 3d gaussian splatting for real-time radiance field rendering.
\newblock \emph{ACM TOG}, 2023.

\bibitem[Kerr et~al.(2024)Kerr, Kim, Wu, Yi, Wang, Goldberg, and Kanazawa]{kerrrobot}
Justin Kerr, Chung~Min Kim, Mingxuan Wu, Brent Yi, Qianqian Wang, Ken Goldberg, and Angjoo Kanazawa.
\newblock Robot see robot do: Imitating articulated object manipulation with monocular 4d reconstruction.
\newblock In \emph{CoRL}, 2024.

\bibitem[Kim et~al.(2024)Kim, Pertsch, Karamcheti, Xiao, Balakrishna, Nair, Rafailov, Foster, Lam, Sanketi, et~al.]{kim2024openvla}
Moo~Jin Kim, Karl Pertsch, Siddharth Karamcheti, Ted Xiao, Ashwin Balakrishna, Suraj Nair, Rafael Rafailov, Ethan Foster, Grace Lam, Pannag Sanketi, et~al.
\newblock Openvla: An open-source vision-language-action model.
\newblock \emph{arXiv:2406.09246}, 2024.

\bibitem[Li et~al.(2022)Li, Li, Sitzmann, Agrawal, and Torralba]{li20223d}
Yunzhu Li, Shuang Li, Vincent Sitzmann, Pulkit Agrawal, and Antonio Torralba.
\newblock 3d neural scene representations for visuomotor control.
\newblock In \emph{CoRL}, 2022.

\bibitem[Liang et~al.(2023)Liang, Huang, Xia, Xu, Hausman, Ichter, Florence, and Zeng]{liang2023code}
Jacky Liang, Wenlong Huang, Fei Xia, Peng Xu, Karol Hausman, Brian Ichter, Pete Florence, and Andy Zeng.
\newblock Code as policies: Language model programs for embodied control.
\newblock In \emph{ICRA}, 2023.

\bibitem[Mildenhall et~al.(2020)Mildenhall, Srinivasan, Tancik, Barron, Ramamoorthi, and Ng]{NeRF}
Ben Mildenhall, Pratul~P Srinivasan, Matthew Tancik, Jonathan~T Barron, Ravi Ramamoorthi, and Ren Ng.
\newblock Nerf: Representing scenes as neural radiance fields for view synthesis.
\newblock In \emph{ECCV}, 2020.

\bibitem[Minniti et~al.(2021)Minniti, Grandia, F{\"a}h, Farshidian, and Hutter]{minniti2021model}
Maria~Vittoria Minniti, Ruben Grandia, Kevin F{\"a}h, Farbod Farshidian, and Marco Hutter.
\newblock Model predictive robot-environment interaction control for mobile manipulation tasks.
\newblock In \emph{ICRA}, 2021.

\bibitem[Mu et~al.(2024)Mu, Zhang, Hu, Wang, Ding, Jin, Wang, Dai, Qiao, and Luo]{mu2024embodiedgpt}
Yao Mu, Qinglong Zhang, Mengkang Hu, Wenhai Wang, Mingyu Ding, Jun Jin, Bin Wang, Jifeng Dai, Yu~Qiao, and Ping Luo.
\newblock Embodiedgpt: Vision-language pre-training via embodied chain of thought.
\newblock In \emph{NeurIPS}, 2024.

\bibitem[Nasiriany et~al.(2024)Nasiriany, Xia, Yu, Xiao, Liang, Dasgupta, Xie, Driess, Wahid, Xu, et~al.]{nasiriany2024pivot}
Soroush Nasiriany, Fei Xia, Wenhao Yu, Ted Xiao, Jacky Liang, Ishita Dasgupta, Annie Xie, Danny Driess, Ayzaan Wahid, Zhuo Xu, et~al.
\newblock Pivot: Iterative visual prompting elicits actionable knowledge for vlms.
\newblock \emph{arXiv:2402.07872}, 2024.

\bibitem[Nubert et~al.(2020)Nubert, K{\"o}hler, Berenz, Allg{\"o}wer, and Trimpe]{nubert2020safe}
Julian Nubert, Johannes K{\"o}hler, Vincent Berenz, Frank Allg{\"o}wer, and Sebastian Trimpe.
\newblock Safe and fast tracking on a robot manipulator: Robust mpc and neural network control.
\newblock \emph{RA-L}, 2020.

\bibitem[Oechsle et~al.(2021)Oechsle, Peng, and Geiger]{oechsle2021unisurf}
Michael Oechsle, Songyou Peng, and Andreas Geiger.
\newblock Unisurf: Unifying neural implicit surfaces and radiance fields for multi-view reconstruction.
\newblock In \emph{ICCV}, 2021.

\bibitem[O'Neill et~al.(2023)O'Neill, Rehman, Gupta, Maddukuri, Gupta, Padalkar, Lee, Pooley, Gupta, Mandlekar, et~al.]{o2023open}
Abby O'Neill, Abdul Rehman, Abhinav Gupta, Abhiram Maddukuri, Abhishek Gupta, Abhishek Padalkar, Abraham Lee, Acorn Pooley, Agrim Gupta, Ajay Mandlekar, et~al.
\newblock Open x-embodiment: Robotic learning datasets and rt-x models.
\newblock \emph{arXiv:2310.08864}, 2023.

\bibitem[Patel et~al.(2024)Patel, Yin, Huang, Garg, Nayyeri, Fei-Fei, Lazebnik, and Li]{patelreal}
Shivansh Patel, Xinchen Yin, Wenlong Huang, Shubham Garg, Hooshang Nayyeri, Li~Fei-Fei, Svetlana Lazebnik, and Yunzhu Li.
\newblock A real-to-sim-to-real approach to robotic manipulation with vlm-generated iterative keypoint rewards.
\newblock In \emph{CoRL Workshop}, 2024.

\bibitem[Peng et~al.(2024)Peng, Lv, Zeng, Chen, Zhao, Sun, Lu, and Shao]{peng2024tiebot}
Weikun Peng, Jun Lv, Yuwei Zeng, Haonan Chen, Siheng Zhao, Jichen Sun, Cewu Lu, and Lin Shao.
\newblock Tiebot: Learning to knot a tie from visual demonstration through a real-to-sim-to-real approach.
\newblock \emph{arXiv:2407.03245}, 2024.

\bibitem[Qureshi et~al.(2024)Qureshi, Garg, Yandun, Held, Kantor, and Silwal]{qureshi2024splatsim}
Mohammad~Nomaan Qureshi, Sparsh Garg, Francisco Yandun, David Held, George Kantor, and Abhisesh Silwal.
\newblock Splatsim: Zero-shot sim2real transfer of rgb manipulation policies using gaussian splatting.
\newblock \emph{arXiv:2409.10161}, 2024.

\bibitem[Ravi et~al.(2024)Ravi, Gabeur, Hu, Hu, Ryali, Ma, Khedr, R{\"a}dle, Rolland, Gustafson, Mintun, Pan, Alwala, Carion, Wu, Girshick, Doll{\'a}r, and Feichtenhofer]{ravi2024sam2}
Nikhila Ravi, Valentin Gabeur, Yuan-Ting Hu, Ronghang Hu, Chaitanya Ryali, Tengyu Ma, Haitham Khedr, Roman R{\"a}dle, Chloe Rolland, Laura Gustafson, Eric Mintun, Junting Pan, Kalyan~Vasudev Alwala, Nicolas Carion, Chao-Yuan Wu, Ross Girshick, Piotr Doll{\'a}r, and Christoph Feichtenhofer.
\newblock Sam 2: Segment anything in images and videos.
\newblock \emph{arXiv:2408.00714}, 2024.

\bibitem[Rubinstein(1999)]{rubinstein1999cross}
Reuven Rubinstein.
\newblock The cross-entropy method for combinatorial and continuous optimization.
\newblock \emph{Methodology and computing in applied probability}, 1999.

\bibitem[Sch\"{o}nberger and Frahm(2016)]{schoenberger2016sfm}
Johannes~Lutz Sch\"{o}nberger and Jan-Michael Frahm.
\newblock Structure-from-motion revisited.
\newblock In \emph{CVPR}, 2016.

\bibitem[Sch\"{o}nberger et~al.(2016)Sch\"{o}nberger, Zheng, Pollefeys, and Frahm]{schoenberger2016mvs}
Johannes~Lutz Sch\"{o}nberger, Enliang Zheng, Marc Pollefeys, and Jan-Michael Frahm.
\newblock Pixelwise view selection for unstructured multi-view stereo.
\newblock In \emph{ECCV}, 2016.

\bibitem[Shen et~al.(2023{\natexlab{a}})Shen, Yang, Yu, Wong, Kaelbling, and Isola]{shen2023distilled}
William Shen, Ge~Yang, Alan Yu, Jansen Wong, Leslie~Pack Kaelbling, and Phillip Isola.
\newblock Distilled feature fields enable few-shot language-guided manipulation.
\newblock In \emph{CoRL}, 2023{\natexlab{a}}.

\bibitem[Shen et~al.(2023{\natexlab{b}})Shen, Yang, Yu, Wong, Kaelbling, and Isola]{shendistilled}
William Shen, Ge~Yang, Alan Yu, Jansen Wong, Leslie~Pack Kaelbling, and Phillip Isola.
\newblock Distilled feature fields enable few-shot language-guided manipulation.
\newblock In \emph{CoRL}, 2023{\natexlab{b}}.

\bibitem[Sun et~al.(2025)]{Sun2024SVR}
Cheng Sun et~al.
\newblock Sparse voxels rasterization: Real-time high-fidelity radiance field rendering.
\newblock \emph{CVPR}, 2025.

\bibitem[Torne et~al.(2024{\natexlab{a}})Torne, Jain, Yuan, Macha, Ankile, Simeonov, Agrawal, and Gupta]{torne2024robot}
Marcel Torne, Arhan Jain, Jiayi Yuan, Vidaaranya Macha, Lars Ankile, Anthony Simeonov, Pulkit Agrawal, and Abhishek Gupta.
\newblock Robot learning with super-linear scaling.
\newblock \emph{arXiv:2412.01770}, 2024{\natexlab{a}}.

\bibitem[Torne et~al.(2024{\natexlab{b}})Torne, Simeonov, Li, Chan, Chen, Gupta, and Agrawal]{torne2024reconciling}
Marcel Torne, Anthony Simeonov, Zechu Li, April Chan, Tao Chen, Abhishek Gupta, and Pulkit Agrawal.
\newblock Reconciling reality through simulation: A real-to-sim-to-real approach for robust manipulation.
\newblock \emph{arXiv:2403.03949}, 2024{\natexlab{b}}.

\bibitem[Wang et~al.(2024{\natexlab{a}})Wang, Zhang, Dong, Fang, and Feng]{wang2024vlm}
Beichen Wang, Juexiao Zhang, Shuwen Dong, Irving Fang, and Chen Feng.
\newblock Vlm see, robot do: Human demo video to robot action plan via vision language model.
\newblock \emph{arXiv:2410.08792}, 2024{\natexlab{a}}.

\bibitem[Wang et~al.(2021)Wang, Liu, Liu, Theobalt, Komura, and Wang]{wang2021neus}
Peng Wang, Lingjie Liu, Yuan Liu, Christian Theobalt, Taku Komura, and Wenping Wang.
\newblock Neus: Learning neural implicit surfaces by volume rendering for multi-view reconstruction.
\newblock \emph{arXiv:2106.10689}, 2021.

\bibitem[Wang et~al.(2023)Wang, Han, Habermann, Daniilidis, Theobalt, and Liu]{wang2023neus2}
Yiming Wang, Qin Han, Marc Habermann, Kostas Daniilidis, Christian Theobalt, and Lingjie Liu.
\newblock Neus2: Fast learning of neural implicit surfaces for multi-view reconstruction.
\newblock In \emph{ICCV}, 2023.

\bibitem[Wang et~al.(2024{\natexlab{b}})Wang, Zhang, Li, Kelestemur, Driggs-Campbell, Wu, Fei-Fei, and Li]{wang2024d}
Yixuan Wang, Mingtong Zhang, Zhuoran Li, Tarik Kelestemur, Katherine~Rose Driggs-Campbell, Jiajun Wu, Li~Fei-Fei, and Yunzhu Li.
\newblock D $^{3}$ fields: Dynamic 3d descriptor fields for zero-shot generalizable rearrangement.
\newblock In \emph{CoRL}, 2024{\natexlab{b}}.

\bibitem[Wu et~al.(2024)Wu, Pan, Wu, Wang, Miao, and Wang]{wu2024rl}
Yuxuan Wu, Lei Pan, Wenhua Wu, Guangming Wang, Yanzi Miao, and Hesheng Wang.
\newblock Rl-gsbridge: 3d gaussian splatting based real2sim2real method for robotic manipulation learning.
\newblock \emph{arXiv:2409.20291}, 2024.

\bibitem[Zawalski et~al.(2024)Zawalski, Chen, Pertsch, Mees, Finn, and Levine]{zawalski2024robotic}
Micha{\l} Zawalski, William Chen, Karl Pertsch, Oier Mees, Chelsea Finn, and Sergey Levine.
\newblock Robotic control via embodied chain-of-thought reasoning.
\newblock \emph{arXiv:2407.08693}, 2024.

\bibitem[Zhao et~al.(2024{\natexlab{a}})Zhao, Chen, Meng, Mao, Song, and Zhang]{zhao2024vlmpc}
Wentao Zhao, Jiaming Chen, Ziyu Meng, Donghui Mao, Ran Song, and Wei Zhang.
\newblock Vlmpc: Vision-language model predictive control for robotic manipulation.
\newblock \emph{arXiv:2407.09829}, 2024{\natexlab{a}}.

\bibitem[Zhao et~al.(2024{\natexlab{b}})Zhao, Cai, Tang, and Wang]{zhao2024imaginenav}
Xinxin Zhao, Wenzhe Cai, Likun Tang, and Teng Wang.
\newblock Imaginenav: Prompting vision-language models as embodied navigator through scene imagination.
\newblock \emph{arXiv:2410.09874}, 2024{\natexlab{b}}.

\end{thebibliography}

\clearpage
\newpage

\appendices
\section{Implementation details}
\subsection{Reconstruction details}
For each scene, we use approximately 200 images for reconstruction. We first run COLMAP~\cite{schoenberger2016sfm, schoenberger2016mvs} to estimate relative camera poses. The chessboard pattern is then used to align these relative poses with the real-world coordinate system. The Gaussian splats are converted into a scene mesh using TSDF volume integration, discarding points with depths larger than a predefined threshold.

To segment objects, we render 100 multi-view images from the reconstructed mesh. A front-view image is then prompted to GPT-4o along with the task instruction to get the names of object that should be segmented (e.g., drum stick in ``playing the drum" and noting in ``press spacebar" since noting need to move other than the robot). The identified object names are passed to Molmo~\cite{deitke2024molmo}, which points to these objects in the front-view image. Finally, SAM2~\cite{ravi2024sam2} is used to segment and track the identified object masks across all multi-view images. Since all images are rendered, we label the mesh vertices that are projected into the 2D masks to segment the 3D object meshes. During reference time, we only render the end effector (i.e., gripper) of the robot to avoid occlusion caused by robot arm. But the motion still obey the arm constraints and we still model arm collision.

There are also some optional post-processing methods to improve the visual quality of the reconstruction. Firstly, when the object is lifted from the table, it leaves a hole in the table since the texture of that part is occluded during scanning. We could automatically use the Gaussian points from the visible table surface to cover the hole, which is shown in all visualizations. Besides, the chessboard used for calibration can be removed by the same technique. Further post-processing could also include pre-scanning the background (e.g., table) and replacing it in the reconstructed digital twin to obtain a better geometry and texture of the background. Please note that we didn't use these techniques in our implementation (except the visualization for illustration purposes), which shows that our framework and VLM is robust to these minor artifacts.

\subsection{Planning details}
At the start of the planning process, we first prompt the VLM with multi-view images of the initial state to facilitate high-level task planning. The VLM decomposes the task into subtasks and determines whether finger movements or gripper rotations are required throughout execution (e.g., pressing a keyboard does not require finger movement or rotation, and the gripper should remain closed). This decision determines the action sampling space for the subsequent model predictive control process.

For each planning step, we render 4 camera view for VLM to choose from, which are front view, left view, right view, and top-down view. VLM is prevented from choose the same view as the previous step to ensure robust spatial perception. 90 actions will be sampled from a Gaussian distribution, results of which are divided into groups of 5 for evaluating the best outcome.

\subsection{Computational details}
Our entire pipeline runs on a single NVIDIA RTX 4090 GPU. Running COLMAP requires approximately 10 minutes, while the reconstruction of Gaussian splats takes 20 minutes. Object segmentation takes about 3 minutes. For planning, each step requires approximately 10 seconds, with 4 seconds for simulation and 6 seconds for VLM prompting.
\section{Experiment Details}

\subsection{Real-world settings}

Our real-world experiments are conducted in a tabletop manipulation environment with a 7-DoF Franka Emika robot arm and a 2F-85 Robotiq gripper. The planned robot trajectory in dthe igital twin are directly copied to the real world by commanding the robot with joint angles. Grasping is controlled by stopping at a specific force threshold.

Our method and VoxPoser*~\cite{huang2023voxposer} use an iPhone camera to scan the static scene. We run MOKA~\cite{fang2024moka} with a wrist RealSense D435 camera (RGB and depth) and OpenVLA~\cite{kim2024openvla} with a thrid-person RealSense D435 camera (RGB) for planning, following their original observation space.

\subsection{Task settings}
\begin{figure}[t]
    \centering
    \includegraphics[width=\linewidth]{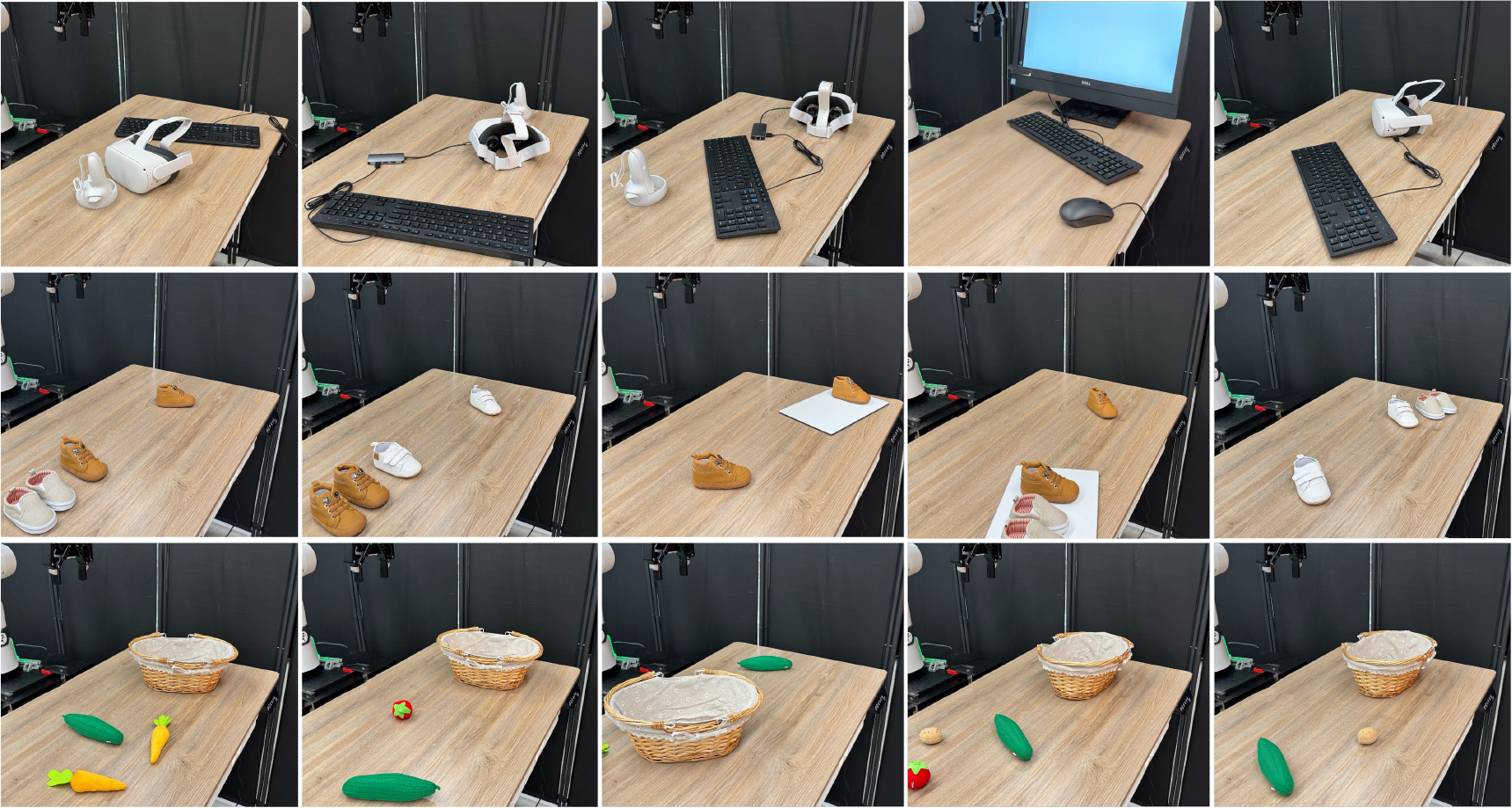}
    \caption{\textbf{Examples of task layouts.} Random object layout and visual distractors for ``press spacebar", ``pair up shoes", and ``cucumber in basket".}
    \label{fig:layout}
\end{figure}

As shown in Figure~\ref{fig:layout}, we randomize the layout of objects and add different visual distractors for five times, resulting in 5 different scenes for each task. Then, we roll out our method and baselines twice on each scene to consider randomness from VLMs. The agent fails when it takes actions that cause irreversible errors or the step budget (30 planning steps) or time budget (5 minutes) is reached. Note that, our method can always replan in digital twins when the VLM decides to. We keep the total planning step (including the replan steps) within the limit of step budget.

\begin{figure*}[!ht]
    \centering
    \includegraphics[width=0.9\linewidth]{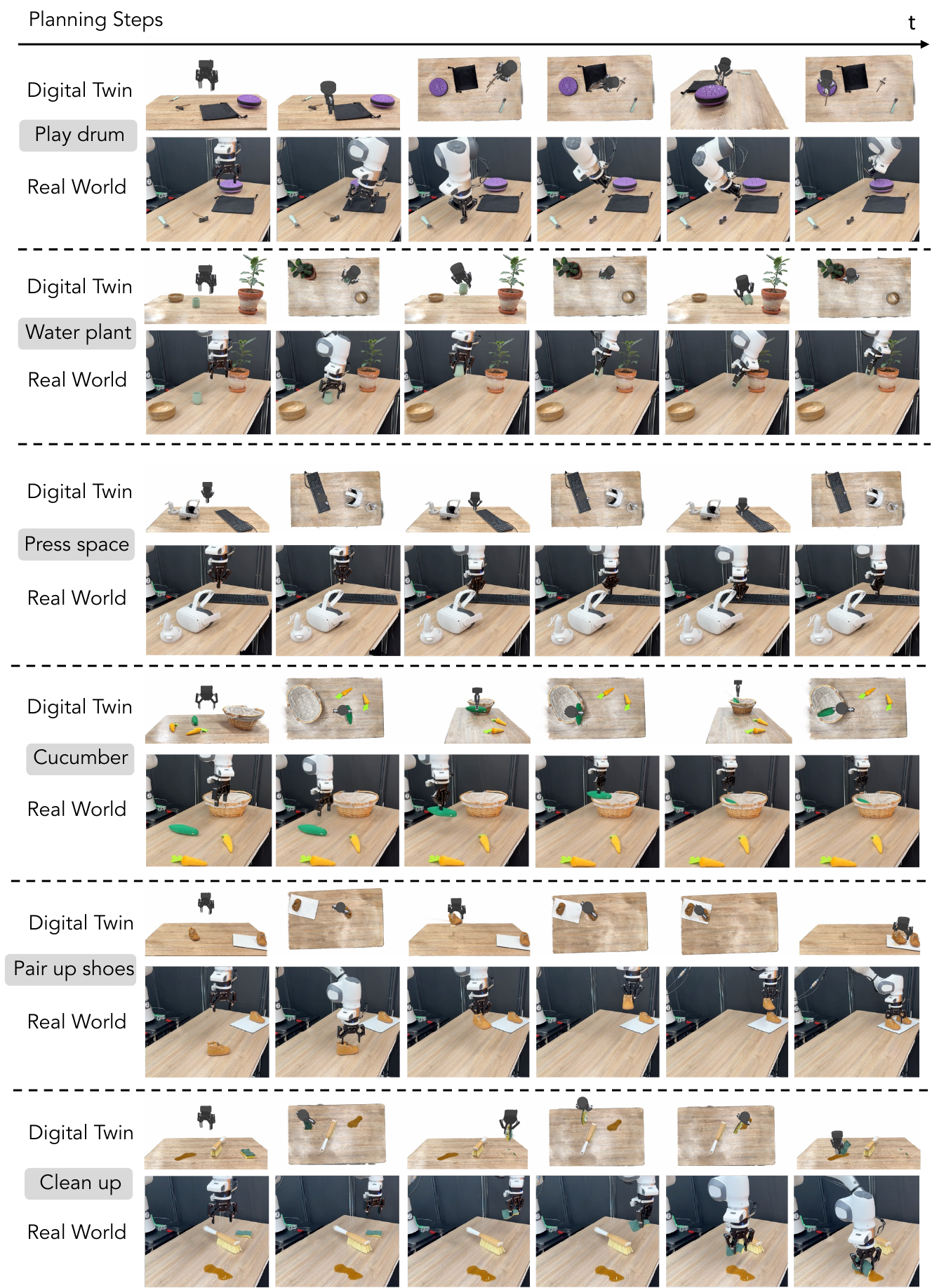}
    \caption{\textbf{Example trajectories.} Example trajectories planned by our framework in both the digital twin and the real world. The camera views in the digital twin are chosen by VLM at each step, some planning steps are omitted for clear visualization.}
    \label{fig:more_results}
\end{figure*}

\begin{figure}[ht]
  \centering
  \includegraphics[width=\linewidth]{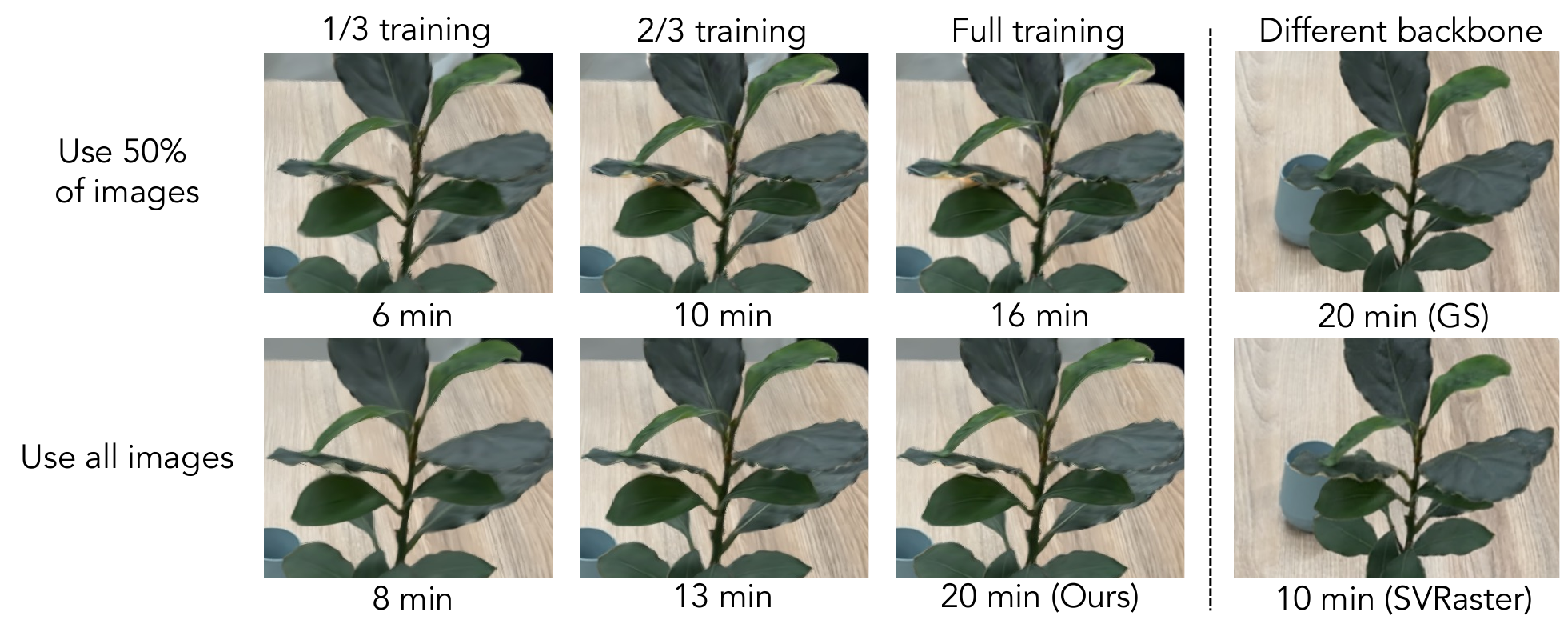}
  \caption{\textbf{Reconstruction quality under different configurations and backbones}. Our framework can easily be sped up by trading off quality and time or incorporating faster techniques.} 
  \label{fig:trade_off}
\end{figure}

\setlength{\tabcolsep}{2.2mm}{
\begin{table}[ht]
    \centering
    \begin{tabular}{lccc}  
    \toprule
    Methods & Water plant & Clean up & Cucumber in basket  \\
    \midrule
    SVRaster~\cite{Sun2024SVR}    &  4/10  & 6/10  & 7/10  \\
    Gaussian Splatting      &  5/10 & 5/10  & 9/10 \\
    \bottomrule
    \end{tabular}
    \caption{\textbf{Success rate with Sparse Voxels Rasterization.} Our method is agnostic and robust to reconstruction methods.}
    \label{tab:svraster}
\end{table}
}

\setlength{\tabcolsep}{2.6mm}{
\begin{table}[ht]
    \centering
    \begin{tabular}{lccc}  
    \toprule
    Configurations & 1/3 training & 2/3 training & Full training  \\
    \midrule
    1/2 images         & 0.3 / 0.4 / 0.8 & 0.3 / 0.3 / 0.8 & 0.2 / 0.4 / 0.9 \\
    Full images      &  0.5 / 0.6 / 0.9 & 0.5 / 0.4 / 0.8 & 0.5 / 0.5 / 0.9 \\
    \bottomrule
    \end{tabular}
    \caption{\textbf{Success rate with different configurations.} We evaluate on Water plant, Clean up, and Cucumber in basket.}
    \label{tab:trade_off}
\end{table}
}

\section{Discussion on Efficiency}
The speed of our 3D reconstruction process depends on both the number of input images and the choice of 3D representation.
While we initially employed Gaussian splatting, our framework is backbone-agnostic and can be seamlessly integrated with faster 3D reconstruction algorithms to improve computational efficiency.
To support this claim, we replaced our 3D backbone with the latest sparse voxel representation~\cite{Sun2024SVR}.
Without any tuning, we reduce reconstruction time by half without sacrificing accuracy (see Tab.~\ref{tab:svraster}).
We anticipate further improvements as 3D reconstruction techniques continue to advance.

Additionally, our framework is robust to the number of input images and reconstruction time (see Fig.~\ref{fig:trade_off} and Tab.~\ref{tab:trade_off}). The performance is comparable even if we reduce the reconstruction time to $\frac{1}{3}$. 

\section{Additional Results}
\subsection{More qualitative results}

We show more example trajectories in both the digital twin and the real world in Figure~\ref{fig:more_results}. Our method is robust to different objects and their layout.

\end{document}